\documentclass{article}
\usepackage{spconf,amsmath,graphicx}
\usepackage{cite}
\usepackage{amsmath,amssymb,amsfonts}
\usepackage{algorithmic}
\usepackage{textcomp}
\usepackage{xcolor}
\usepackage{color}
\usepackage{colortbl}
\usepackage{rotating}
\usepackage{multirow}
\usepackage{booktabs}
\usepackage{array}
\usepackage[colorlinks,
            linkcolor=red,
            anchorcolor=blue,
            citecolor=green]{hyperref}

\def\ie{\emph{i.e.}}
\def\eg{\emph{e.g.}}
\def\etc{\emph{etc}}

\title{DEPTH-COOPERATED TRIMODAL NETWORK FOR VIDEO\\SALIENT OBJECT DETECTION}
\name{Yukang Lu, Dingyao Min, Keren Fu\sthanks{This research is partly supported by NSFC (62176169, 62176170), and SCU-Luzhou Municipal Peoples Government Strategic Cooperation Project (2020CDLZ-10). Corresponding author: Keren Fu (fkrsuper@scu.edu.cn).}, Qijun Zhao \vspace{-6pt}}
\address{College of Computer Science, Sichuan University\\
   National Key Laboratory of Fundamental Science on Synthetic Vision, Sichuan University}
\begin{document}\sloppy
\topmargin=0mm
%
\maketitle
\begin{abstract}
   Depth can provide useful geographical cues for salient object detection (SOD), and has been proven helpful in recent RGB-D SOD methods. However, existing video salient object detection (VSOD) methods only utilize spatiotemporal information and seldom exploit depth information for detection. In this paper, we propose a depth-cooperated trimodal network, called DCTNet for VSOD, which is a pioneering work to incorporate depth information to assist VSOD. To this end, we first generate depth from RGB frames, and then propose an approach to treat the three modalities unequally. Specifically, a multi-modal attention module (MAM) is designed to model multi-modal long-range dependencies between the main modality (RGB) and the two auxiliary modalities (depth, optical flow). We also introduce a refinement fusion module (RFM) to suppress noises in each modality and select useful information dynamically for further feature refinement. Lastly, a progressive fusion strategy is adopted after the refined features to achieve final cross-modal fusion. Experiments on five benchmark datasets demonstrate the superiority of our depth-cooperated model against 12 state-of-the-art methods, and the necessity of depth is also validated. 
\end{abstract}
\vspace{-5pt}
\section{INTRODUCTION} \label{sec:intro}
\vspace{-5pt}

Video salient object detection (VSOD) aims to locate the most attention-grabbing objects in a video clip, and plays an important role in many down-stream vision tasks, such as video segmentation, tracking, autonomous driving, \etc{}. 
Nowadays, with the rapid development of deep learning, CNN-based methods have dominated the VSOD field. Emerging methods try to exploit both spatial and temporal information for detection and can be roughly divided into 3D convolution-based methods \cite{le2018video}, recurrent neural networks-based methods \cite{fgrne, pdbm, ssav}, and optical flow-based methods \cite{mgan, tenet, fsnet}. 
Meanwhile, it is believed that depth maps can provide useful geographical cues to improve performance for the SOD task, especially when handing challenging scenarios, \eg{}, in low contrast or cluttered background. As a result, depth has been proven very helpful in recent RGB-D SOD models \cite{fu2020jl,fan2020bbs,zhou2021specificity}.

Inspired by the advances \cite{mgan,tenet,fsnet, fu2020jl,fan2020bbs,zhou2021specificity} in the above two areas, in this paper we attempt to exploit depth information for the VSOD task, which still remains \emph{under-explored} as \emph{existing VSOD methods seldom incorporate depth to assist detection, not like that in the RGB-D SOD task}.
Note that in the related zero-shot video object segmentation (ZVOS) field, \cite{multi} segments objects using multiple sources including depth maps. However, that work is related to another area and also a multi-stage method, which is sophisticated to apply in practice.

\begin{figure}
\includegraphics[width=.48\textwidth]{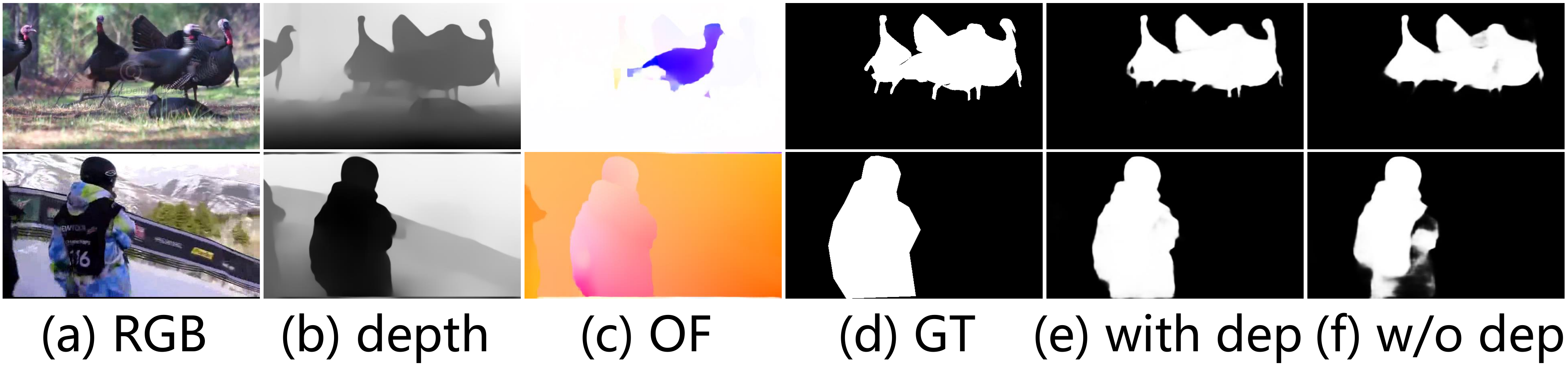}\vspace{-0.4cm}
\caption{\small Effectiveness of leveraging depth to assist VSOD. OF denotes optical flow, and GT represents ground truth. Column (e) and (f) are predictions from our full model (with depth) and its variant (without depth, the A1 model in Sec.\ref{Ablation}), respectively.}
\label{fig_nodepth}
\vspace{-0.5cm}
\end{figure}

\begin{figure*}
\includegraphics[width=.98\textwidth]{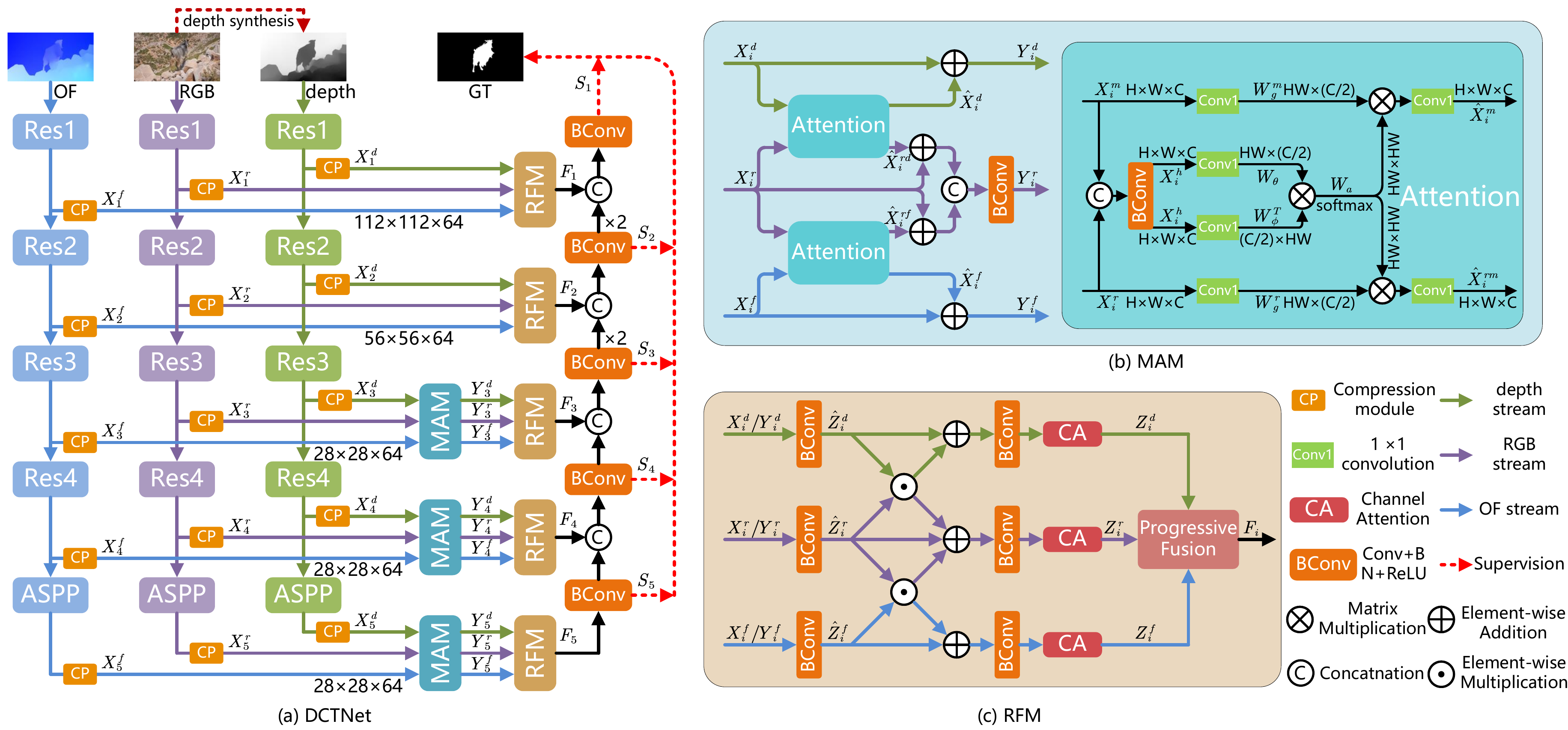} 
\vspace{-0.3cm}
\caption{\small Overview of DCTNet. (a) shows the big picture. (b) and (c) show the details of MAM and RFM, respectively.}
\label{fig_blockdiagram}
\vspace{-0.4cm}
\end{figure*}

To explore the contribution of depth in VSOD, we propose a novel network called depth-cooperated trimodal network (DCTNet), leveraging depth as assistance for VSOD.
Since current VSOD datasets do not provide depth maps, 
we first conduct depth prediction by using an off-the-shelf monocular depth estimator. The obtained depth information, together with optical flow (OF) and RGB images, form a trimodal input for the subsequent DCTNet.    
While most previous multi-modal methods \cite{fsnet,fu2020jl,zhou2021specificity} treat modalities equally, neglecting their underlying differences, inspired by \cite{mgan}, we propose to treat them unequally with one main modality (RGB) and two auxiliary modalities (depth, OF). As a result, a multi-modal attention module (MAM) and refinement fusion module (RFM) are designed. The former helps to enhance features by modeling multi-modal long-range dependencies, whereas the latter suppresses individual noises, selects features for refinement, and finally fuses them in a progressive way.
Fig.\:\ref{fig_nodepth} shows some visual results, where our method cooperated with depth can obtain better performance
compared to its variant that does not leverage depth.

Our main contributions can be summarized as follows:
\vspace{-0.3cm}
\begin{itemize}
    \setlength{\itemsep}{0pt}
    \setlength{\parsep}{0pt}
    \setlength{\parskip}{0pt}
    \setlength{\topsep}{0pt}
    \item We propose a depth-cooperated trimodal network (DCTNet), which is a pioneering model leveraging depth information as assistance for VSOD. 
    \item Different from previous \cite{fsnet,fu2020jl,zhou2021specificity}, we propose to treat different modalities unequally during cross-modal fusion, yielding one main modality (RGB) and two auxiliary modalities (depth, OF). Guided by this idea, we propose two novel designs, \emph{i.e.}, multi-modal attention module (MAM) and refinement fusion module (RFM). 
    \item Extensive experiments show the superiority of our DCTNet against 12 state-of-the-art VSOD methods, and the benefit of incorporating depth is also validated. 
\end{itemize}

\vspace{-6pt}
\section{METHODOLOGY} \label{sec:method}
\vspace{-6pt}

The overview of DCTNet is shown in Fig.\:\ref{fig_blockdiagram}, which is characterized by a three-stream encoder-decoder architecture. To obtain scene depth information, we employ DPT \cite{dpt} to generate synthetic depth maps from individual frames. OF maps are rendered by RAFT \cite{raft}. Each stream of the encoder is based on ResNet-34 \cite{resnet}, and following \cite{mgan}, the ASPP (atrous spatial pyramid pooling \cite{deeplab}) module is attached to the last layer.
Let the extracted five-level RGB features be $(X_{i}^{r})_{i=1}^5$, depth features be $(X_{i}^{d})_{i=1}^5$, and OF features be $(X_{i}^{f})_{i=1}^5$. Their channel numbers are aligned to 64 via the compression module (CP) for computational simplicity. CP is actually realized by the $BConv$ operation mentioned later. Next, we treat RGB as the main modality, and depth and OF as two auxiliary modalities. Their features are handled unequally in multi-modal attention module (MAM) and refinement fusion module (RFM). The former helps to enhance features, and the latter refines and fuses features. Details are described below.

\vspace{-1pt}
\textbf{Multi-modal Attention Module (MAM).} 
Long-range information is well-known for enhancing features, which can be captured by the non-local (NL) module \cite{wang2018non}. Considering that NL only explores long-range information of one modality, we design MAM based on NL to model multi-modal long-range dependencies for enhancing trimodal features. Due to computational limitation, MAM only works on the higher three hierarchies $X_{i}^{p}\in \mathbb{R}^{H \times W \times C}(p \in\{r, d, f\}, i=3,4,5)$, where $H$,$W$,$C$ refer to the height, width, and channel number, respectively. Fig.\:\ref{fig_blockdiagram}(b) shows the inner structure of MAM, where the main RGB feature are interacted with either auxiliary features $X_{i}^{m}(m \in\{d, f\})$ in the attention block. Specifically,
we first combine the main RGB features and auxiliary features into multi-modal features, as $X_{i}^{h} = BConv([X_{i}^{r},X_{i}^{m}])$, where $[\cdot,\cdot]$ denotes channel concatenation, and $BConv$ comprises convolution,  BatchNorm and ReLU. Following \cite{wang2018non}, $1 \times 1$ convolutions (Conv1 in Fig.\:\ref{fig_blockdiagram}(b)) are used to embed $X_{i}^{h}$, yielding $W_{\theta}$ and $W_{\phi}$. Then pairwise relationships of multi-modal information are captured by ``query-key matching'' as follows:
\setlength{\abovedisplayskip}{4pt}
\setlength{\belowdisplayskip}{4pt}
\begin{align}
    W_{a} &= softmax(W_{\theta} \otimes W_{\phi}^{T})
\end{align}
where $\otimes$ is matrix multiplication and $W_{a}\in \mathbb{R}^{HW \times H W}$.
Further, multi-modal long-range dependencies are modeled by transmitting the multi-modal affinity matrix $W_{a}$ to the main RGB and auxiliary feature embedding, denoted as $W_{g}^{r}$ and $W_{g}^{m}$ in Fig.\:\ref{fig_blockdiagram}(b):
\begin{align}
    \hat{X}_{i}^{rm} =W_{a}\otimes W_{g}^{r},~\hat{X}_{i}^{m} =W_{a}\otimes W_{g}^{m}  
\end{align}
where $\hat{X}_{i}^{rm}$ and $\hat{X}_{i}^{m}$ mean cross-modal long-range attended features (note Conv1 after $\otimes$ is omitted, and $m \in\{d, f\}$).

For either auxiliary modality, the enhanced features are obtained by adding $\hat{X}_{i}^{m}$ back to the original features, namely $Y_{i}^{m} = X_{i}^{m} + \hat{X}_{i}^{m}$. 
Meanwhile, for the main RGB modality, we further concatenate the results assisted by the two auxiliary modalities, and then transform and aggregate the features using $BConv$, yielding $Y_{i}^{r}$ as shown in Fig.\:\ref{fig_blockdiagram}(b) left. In this way, all three modality-aware features are fully enhanced by propagating cross-modal long-range information. 

Note in this trimodal task, we find that cross-modal long-range information is more powerful for feature enhancement than self long-range one. Meanwhile, we design RGB to be the main modality because it is more stable/reliable and serves as basic for saliency detection \cite{mgan, fan2020bbs}. Ablation experiments in Sec.\ref{Ablation} will validate the above design ideas of MAM.

\vspace{-2pt}
\textbf{Refinement Fusion Module (RFM).} 
There are noises in each modality-aware features, which are detrimental to accurate saliency maps. Directly fusing trimodal features together may inevitably lead to contamination. RFM is proposed to address this issue, which is applied on all hierarchies as shown in Fig.\:\ref{fig_blockdiagram}(a). 
As illustrated in Fig.\:\ref{fig_blockdiagram}(c), trimodal
features are first adapted by $BConv$ for subsequent processing, yielding aligned features $\hat{Z}_{i}^{p}(p \in\{r, d, f\}, i=1,...,5)$.
Since useful information and much less background noises are contained in shared areas between main and auxiliary modalities. To refine features, we extract common information between the main modality (RGB) and either auxiliary modality (depth/OF) through element-wise multiplication. Such common information is then added back to the original features, in order to repress individual background noises and meanwhile purify features. To achieve better refinement, we also employ channel attention \cite{hu2018squeeze} to dynamically select features helpful for SOD. The above process is summarized as below:
\begin{align}
    Z_{i}^{r} &= CA(BConv(\hat{Z}_{i}^{r} \odot \hat{Z}_{i}^{d} + \hat{Z}_{i}^{r} \odot \hat{Z}_{i}^{f} + \hat{Z}_{i}^{r})) \\
    Z_{i}^{m} &= CA(BConv(\hat{Z}_{i}^{m} \odot \hat{Z}_{i}^{r} + \hat{Z}_{i}^{m})),~m \in\{d, f\} 
\end{align} 
where $CA$ denotes the channel attention operation, and $\odot$ is element-wise multiplication. Then the refined trimodal features are combined together to get high-quality fused features. 

\vspace{-1.5pt}
To adaptively adjust assistant influence of the two auxiliary modalities, we first fuse their features together, and then combine the result with the main RGB features to get final fused features, which is formulated as:
\begin{align}
    F_{i} = Bconv([Z_{i}^{r},Bconv([Z_{i}^{d},Z_{i}^{f}])])
\end{align} 
Such a progressive fusion strategy is shown in Fig.\:\ref{fig_blockdiagram}(c) as a simplified block.
Ablation experiments in Sec.\ref{Ablation} show it gives better performance than equally concatenating three features. 
Finally, hierarchical fused features are $\times 2$ bilinear interpolated to match the lower layer optionally, and they are concatenated and processed by $BConv$, composing a typical U-Net decoder \cite{2015unet} to achieve final saliency prediction.


\vspace{-1.5pt}
\textbf{Supervision.}  
We adopt a combination of widely used binary cross entropy loss and intersection-over-union loss \cite{iouloss} for training DCTNet. The total loss is formulated as: $
    l_{total} = \sum_{i = 1}^{5} ({1}/{2^{i-1}}) l(S_{i},G)
$. $S_{i}$ is the output from $i$-th layer of the decoder. $G$ represents the ground truth. $l(S_{i},G)$ denotes the combined loss. Note we assign higher-level loss the lower weight (\emph{i.e.}, ${1}/{2^{i-1}}$) due to its larger error. During inference, we take $S_{1}$ as the final saliency prediction.

\vspace{-6pt}
\section{EXPERIMENTS} \label{sec:Experiments}
\vspace{-6pt}

\begin{table*}[t!]
    \renewcommand{\arraystretch}{1.0}
    \caption{\small Quantitative comparison with state-of-the-art VSOD methods on 5 benchmark datasets. The best and second best results are shown in \textcolor{red}{red} and \textcolor{blue}{blue} respectively. $\uparrow$/$\downarrow$ denotes that the larger/smaller value is better. Symbol `**' means that results are not available.
    }\label{tab_sota}
    \vspace{0.1cm}
  \centering
    \footnotesize
    \setlength{\tabcolsep}{1.25mm}
    
    \begin{tabular}{c|c|ccc|ccc|ccc|ccc|ccc}
    \hline
    \multirow{2}{*}{Methods}&\multirow{2}{*}{Year}&\multicolumn{3}{c|}{DAVIS\cite{davis}}&
    \multicolumn{3}{c|}{DAVSOD\cite{ssav}}&\multicolumn{3}{c|}{FBMS\cite{fbms}}&\multicolumn{3}{c|}{SegV2\cite{segv2}}&\multicolumn{3}{c}{VOS\cite{vos}}\cr\cline{3-17}
    &&$F_{\beta}^{\textrm{max}}\uparrow$&$S_\alpha\uparrow$&$M\downarrow$&$F_{\beta}^{\textrm{max}}\uparrow$&$S_\alpha\uparrow$&$M\downarrow$&$F_{\beta}^{\textrm{max}}\uparrow$&$S_\alpha\uparrow$&$M\downarrow$&$F_{\beta}^{\textrm{max}}\uparrow$&$S_\alpha\uparrow$&$M\downarrow$&$F_{\beta}^{\textrm{max}}\uparrow$&$S_\alpha\uparrow$&$M\downarrow$\cr
    \hline
    
    MSTM\cite{mstm}&CVPR'16&0.395&0.566&0.174&0.347&0.530&0.214&0.500&0.613&0.177&0.526&0.643&0.114&0.567&0.657&0.144 \cr
    STBP\cite{stbp}&TIP'16&0.485&0.651&0.105&0.408&0.563&0.165&0.595&0.627&0.152&0.640&0.735&0.061&0.526&0.576&0.163 \cr
    SFLR\cite{sflr}&TIP'17&0.698&0.771&0.060&0.482&0.622&0.136&0.660&0.699&0.117&0.745&0.804&0.037&0.546&0.624&0.145\cr
    SCOM\cite{scom}&TIP'18&0.746&0.814&0.055&0.473&0.603&0.219&0.797&0.794&0.079&0.764&0.815&0.030&0.690&0.712&0.162 \cr
    SCNN\cite{scnn}&TCSVT'18&0.679&0.761&0.077&0.494&0.680&0.127&0.762&0.794&0.095&**&**&**&0.609&0.704&0.109 \cr
    FGRNE\cite{fgrne}&CVPR18&0.783&0.838&0.043&0.589&0.701&0.095&0.767&0.809&0.088&0.694&0.770&0.035&0.669&0.715&0.097 \cr
    PDBM\cite{pdbm}&ECCV'18&0.855&0.882&0.028&0.572&0.698&0.116&0.821&0.851&0.064&0.808&0.864&0.024&0.742&0.817&0.078 \cr
    SSAV\cite{ssav}&CVPR'19&0.861&0.893&0.028&0.603&0.724&0.092&0.865&0.879&0.040&0.798&0.851&\textcolor{blue}{0.023}&0.742&0.819&0.074 \cr
    MGAN\cite{mgan}&ICCV'19&0.893&0.913&0.022&0.662&0.757&0.079&\textcolor{blue}{0.909}&\textcolor{red}{0.912}&\textcolor{blue}{0.026}&\textcolor{red}{0.840}&\textcolor{red}{0.895}&0.024&0.743&0.807&0.069 \cr
    PCSA\cite{pcsa}&AAAI'20&0.880&0.902&0.022&0.656&0.741&0.086&0.837&0.868&0.040&\textcolor{blue}{0.811}&0.866&0.024&\textcolor{blue}{0.747}&\textcolor{blue}{0.828}&\textcolor{blue}{0.065} \cr
    TENet\cite{tenet}&ECCV'20&0.894&0.905&0.021&0.648&0.753&0.078&0.887&0.910&0.027&**&**&**&**&**&** \cr
    FSNet\cite{fsnet}&ICCV'21&\textcolor{blue}{0.907}&\textcolor{blue}{0.920}&\textcolor{blue}{0.020}&\textcolor{blue}{0.685}&\textcolor{blue}{0.773}&\textcolor{blue}{0.072}&0.888&0.890&0.041&0.806&0.870&0.025&0.659&0.703&0.103 \cr
    DCTNet(Ours)& 2022 &\textcolor{red}{0.912}&\textcolor{red}{0.922}&\textcolor{red}{0.015}&\textcolor{red}{0.728}&\textcolor{red}{0.797}&\textcolor{red}{0.061}&\textcolor{red}{0.913}&\textcolor{blue}{0.911}&\textcolor{red}{0.025}&\textcolor{red}{0.840}&\textcolor{blue}{0.889}&\textcolor{red}{0.019}&\textcolor{red}{0.793}&\textcolor{red}{0.846}&\textcolor{red}{0.051} \cr
    \hline
    \end{tabular}
\vspace{-0.4cm}
\end{table*}
\vspace{-5pt}
\subsection{Datasets, Metrics and Implementation Details}\label{sec41}
\vspace{-7pt}
We conduct experiments on five VSOD benchmark datasets, \ie{}, DAVIS\cite{davis}, DAVSOD\cite{ssav}, FBMS\cite{fbms}, SegTrack-V2\cite{segv2} and VOS\cite{vos}. For performance evaluation, we adopt three widely used evaluation metrics, including: S-measure ($S_\alpha$)\cite{smeasure}, maximum F-measure ($F_{\beta}^{\textrm{max}}$)\cite{maxf}  and MAE ($M$)\cite{mae}. 
Our DCTNet is implemented 
by PyTorch.
Following a similar training strategy as \cite{mgan,fsnet}, we first pre-train the depth stream on synthetic depth maps and the OF stream on rendered OF images. We also employ an image saliency dataset DUTS\cite{duts} to pre-train the RGB stream. Our entire trimodal network is finally fine-tuned on the training sets of DAVIS, DAVSOD, and FBMS, which contain 30, 29 and 61 clips, respectively. Trimodal inputs of RGB, depth and OF are resized to $448 \times 448 $ with batch size 8. We adopt the SGD algorithm to optimize. The initial learning rates of backbones and other parts are set to 1e-4 and 1e-3, respectively. Data augmentation including random flipping, random cropping are utilized.

\begin{figure}[t]
\includegraphics[width=.48\textwidth]{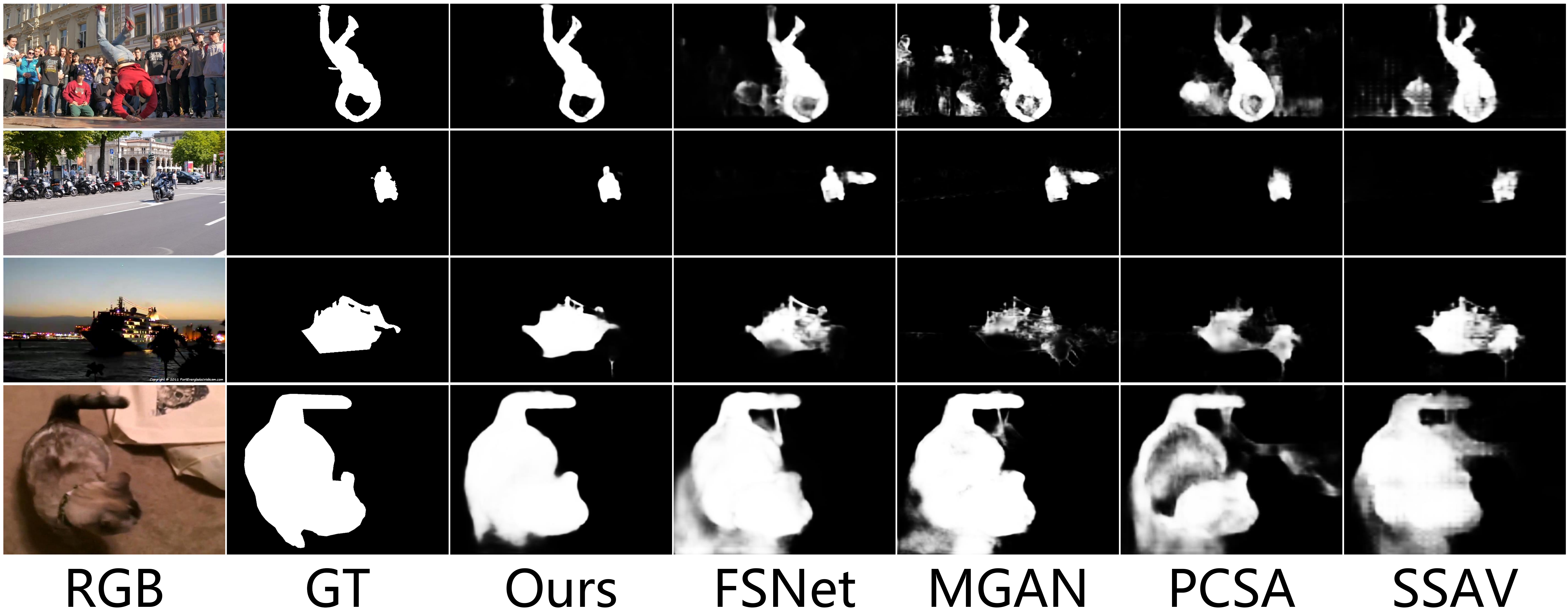} 
\vspace{-0.8cm}
\caption{\small Qualitative comparison of our method and SOTA methods.}
\label{visual}
\vspace{-0.4cm}
\end{figure}

\vspace{-10pt}
\subsection{Comparison to State-of-the-Arts}\label{sec42}
\vspace{-7pt}

To demonstrate the effectiveness of our proposed method, we conduct comparison with 12 state-of-the-art (SOTA) VSOD methods including 4 traditional methods\cite{mstm,stbp,sflr,scom} and 8 deep learning-based methods\cite{scnn,fgrne,pdbm,ssav,mgan,pcsa,tenet,fsnet}.
Quantitative results in Table\:\ref{tab_sota} show that our method achieves superior performance against almost all these models without any post-processing.
For qualitative evaluation, we show visual comparison of DCTNet with SOTA methods in Fig.\:\ref{visual}. 
It can be seen that our method is able to generate more accurate and complete saliency maps, effectively handling difficult scenarios, such as cluttered background (row 1), moving background (row 2), low-illumination scene (row 3) and low contrast (row 4).

\vspace{-10pt}
\subsection{Ablation Study}\label{Ablation}
\vspace{-7pt}

\noindent \textbf{Effectiveness of incorporating depth.} The main contribution of our method is introducing depth information for VSOD. To validate its rationality, we change the two proposed modules by removing depth-related parts, and also exclude the depth stream. This retained two-stream variant is denoted as A1, which is re-trained and tested using RGB and OF. The results on DAVIS, DAVSOD and VOS are shown in Table\:\ref{tab_ab1}, which shows incorporating depth does boost performance by notable margins.

\begin{table}[t!]

  \renewcommand{\arraystretch}{1.0}
  \vspace{-0.3cm}
    \caption{Ablation results. The best are highlighted in \bf{bold}.}\label{tab_ab1}
    \vspace{0.1cm}
   \centering
    \footnotesize
    \setlength{\tabcolsep}{2.35mm}
    \begin{tabular}{c|cc|cc|cc}
    \hline
    \multirow{2}{*}{}&\multicolumn{2}{c|}{DAVIS}&
    \multicolumn{2}{c|}{DAVSOD}&\multicolumn{2}{c}{VOS}\cr\cline{2-7}
    &$F_{\beta}^{\textrm{max}}\uparrow$&$S_\alpha\uparrow$&$F_{\beta}^{\textrm{max}}\uparrow$&$S_\alpha\uparrow$&$F_{\beta}^{\textrm{max}}\uparrow$&$S_\alpha\uparrow$\cr
    \hline
    A1&0.901&0.916&0.709&0.784&0.782&0.835\cr
    \hline
    B1&0.898&0.916&0.705&0.778&0.772&0.821\cr 
    B2&0.899&0.915&0.703&0.781&0.773&0.828\cr 
    \hline
    C1&0.909&0.921&0.704&0.784&0.779&0.835\cr 
    C2& \bf0.912& \bf0.922&0.711&0.788&0.785&0.838 \cr
    C3&0.906&0.921&0.715&0.790&0.785&0.838\cr
    C4&0.906&0.920&0.724&0.795&0.791&0.841 \cr
    \hline

     Ours&\bf 0.912& \bf 0.922&\bf 0.728&\bf0.797&\bf0.793&\bf0.846\cr
    \hline
    \end{tabular}
    \vspace{-0.4cm}
\end{table}

\noindent \textbf{Choice of the main modality.} Since we treat different modalities unequally, and take RGB as the main modality, we conduct experiments to validate this choice by choosing depth/OF as the main one, yielding variants B1 and B2. As shown in Table\:\ref{tab_ab1}, both of them achieve inferior results compared to our full model.

\noindent \textbf{Effectiveness of the proposed modules.} To validate the proposed modules MAM and RFM, we conduct experiments by removing or replacing them from our full model. C1 means the variant without MAM, and C2 replaces MAM with three original self non-local modules. By comparing C1/C2 with our full model, it can be seen that our MAM brings higher boost attributed to multi-modal long-range dependencies. We also replace RFM by simple concatenation without refinement, yielding C3. To further show the usefulness of progressive fusion in RFM, we directly concatenate the refined features in RFM instead, yielding C4. Results in Table\:\ref{tab_ab1} prove our inner designs in RFM, and C4 is also better than C3.

\noindent \textbf{Impact of synthesized depth maps.} 
As aforementioned, we employ depth estimator DPT \cite{dpt} for DCTNet. Intuitively, the quality of synthesized depth should have certain impact on the results. To study this, we replace it with  Monodepth2\cite{monodepth2} and MegaDepth\cite{megadepth}, whose depth prediction accuracy is inferior to DPT, and models are re-trained likewise. Table\:\ref{tab_ab2} shows the obtained results, where one can see higher quality depth maps are conducive to the proposed DCTNet. But even the worst Monodepth2 is better than A1 without depth. 


\begin{table}[t!]

  \renewcommand{\arraystretch}{1.1}
  \vspace{-0.3cm}
    \caption{Results of different depth estimators in our DCTNet.}\label{tab_ab2}
    \vspace{0.1cm}
   \centering
    \footnotesize
    \setlength{\tabcolsep}{1.6mm}
    \begin{tabular}{c|cc|cc|cc}
    \hline
    \multirow{2}{*}{Depth estimator}&\multicolumn{2}{c|}{DAVIS}&
    \multicolumn{2}{c|}{DAVSOD}&\multicolumn{2}{c}{VOS}\cr\cline{2-7}
    &$F_{\beta}^{\textrm{max}}\uparrow$&$S_\alpha\uparrow$&$F_{\beta}^{\textrm{max}}\uparrow$&$S_\alpha\uparrow$&$F_{\beta}^{\textrm{max}}\uparrow$&$S_\alpha\uparrow$\cr
    \hline
    Monodepth2\cite{monodepth2}&0.905&0.916&0.716&0.788&0.782&0.834\cr
    MegaDepth\cite{megadepth}&0.909&0.918&0.718&0.788&0.787&0.837\cr 
    DPT\cite{dpt}&\bf 0.912& \bf 0.922&\bf 0.728&\bf0.797&\bf0.793&\bf0.846\cr
    \hline
    \end{tabular}
    \vspace{-0.4cm}
\end{table}

\begin{table}[t!]

  \renewcommand{\arraystretch}{1.12}
  \vspace{-0.2cm}
    \caption{Results on our VSOD dataset with realistic depth.}\label{tab_RDVSOD}
    \vspace{0.1cm}
   \centering
    \footnotesize
    \begin{tabular}{p{1.1cm}<{\centering}|p{1.3cm}<{\centering}|p{1.3cm}<{\centering}|p{1.2cm}<{\centering}|p{1.2cm}<{\centering}}
    \hline
      Metric &MGAN \cite{mgan} & FSNet \cite{fsnet} & A1  & Ours \\
    \hline
    $F_{\beta}^{\textrm{max}}\uparrow$ &0.824&0.803& 0.824 &\bf 0.841\\
    $S_\alpha\uparrow$                &0.859&0.842&0.858&\bf 0.866\\
    $M\downarrow$                     &0.044&0.049&0.043&\bf 0.041\\
    
    \hline
    \end{tabular}
    \vspace{-0.5cm}
\end{table}

\vspace{-7pt}
\subsection{On VSOD with Realistic Depth}\label{Dataset}
\vspace{-7pt}
Nevertheless, synthetic depth maps are often limited and inaccurate to reflect real-world, compared to those acquired by depth sensors. Therefore, we believe that realistic depth information can also assist VSOD since depth sensors and RGB-D videos are becoming popular.
To this end, we construct an RGB-D video dataset for SOD (will be released soon), comprising data collected from existing public RGB-D video datasets. We densely annotate saliency masks of target objects, resulting in $\sim$4087 frames in total. Multiple challenging scenarios are also included. We conduct evaluation on this dataset, including two OF-based models \cite{mgan,fsnet}, and our A1 in Sec.\ref{Ablation} and full DCTNet. Results in Table\:\ref{tab_RDVSOD} show that DCTNet performs encouragingly on this dataset with realistic depth being fed, shedding light on incorporating depth and motion together for the SOD task.


\vspace{-5pt}
\section{CONCLUSION}\label{sec4}
\vspace{-5pt}

In this paper, we propose a novel depth-cooperated trimodal network (DCTNet) for VSOD, which leverages generated depth as intermediate information for trimodal input. Our model treats the three modalities unequally, with RGB being the main modality. Two new modules, namely multi-modal attention module (MAM) and refinement fusion module (RFM), are proposed for feature enhancement and refinement.  
Extensive experiments have validated our method, especially incorporating depth for VSOD. As RGB-D videos are becoming popular, in the future we may extend and improve our model to deal with video saliency with realistic depth, which we consider as the RGB-D VSOD problem.



\vfill\pagebreak

\small
\bibliographystyle{IEEEbib}
\bibliography{refs}

\begin{thebibliography}{10}

\bibitem{le2018video}
T.-N. Le and A.~Sugimoto,
\newblock ``Video salient object detection using spatiotemporal deep
  features,''
\newblock {\em IEEE TIP}, vol. 27, no. 10, pp. 5002--5015, 2018.

\bibitem{fgrne}
G.~Li, Y.~Xie, T.~Wei, K.~Wang, and L.~Lin,
\newblock ``Flow guided recurrent neural encoder for video salient object
  detection,''
\newblock in {\em CVPR}, 2018, pp. 3243--3252.

\bibitem{pdbm}
H.~Song, W.~Wang, S.~Zhao, J.~Shen, and K.-M. Lam,
\newblock ``Pyramid dilated deeper convlstm for video salient object
  detection,''
\newblock in {\em ECCV}, 2018, pp. 715--731.

\bibitem{ssav}
D.-P. Fan, W.~Wang, M.-M. Cheng, and J.~Shen,
\newblock ``Shifting more attention to video salient object detection,''
\newblock in {\em CVPR}, 2019, pp. 8554--8564.

\bibitem{mgan}
H.~Li, G.~Chen, G.~Li, and Y.~Yizhou,
\newblock ``Motion guided attention for video salient object detection,''
\newblock in {\em ICCV}, 2019, pp. 7274--7283.

\bibitem{tenet}
S.~Ren, C.~Han, X.~Yang, G.~Han, and S.~He,
\newblock ``Tenet: Triple excitation network for video salient object
  detection,''
\newblock in {\em ECCV}, 2020, pp. 212--228.

\bibitem{fsnet}
G.-P. Ji, K.~Fu, Z.~Wu, D.-P. Fan, J.~Shen, and L.~Shao,
\newblock ``Full-duplex strategy for video object segmentation,''
\newblock in {\em ICCV}, 2021, pp. 4922--4933.

\bibitem{fu2020jl}
K.~Fu, D.-P. Fan, G.-P. Ji, and Q.~Zhao,
\newblock ``Jl-dcf: Joint learning and densely-cooperative fusion framework for
  rgb-d salient object detection,''
\newblock in {\em CVPR}, 2020, pp. 3052--3062.

\bibitem{fan2020bbs}
D.-P. Fan, Y.~Zhai, A.~Borji, J.~Yang, and L.~Shao,
\newblock ``Bbs-net: Rgb-d salient object detection with a bifurcated backbone
  strategy network,''
\newblock in {\em ECCV}, 2020, pp. 275--292.

\bibitem{zhou2021specificity}
T.~Zhou, H.~Fu, G.~Chen, Y.~Zhou, D.-P. Fan, and L.~Shao,
\newblock ``Specificity-preserving rgb-d saliency detection,''
\newblock in {\em ICCV}, 2021, pp. 4681--4691.

\bibitem{multi}
X.~Zhao, Y.~Pang, J.~Yang, L.~Zhang, and H.~Lu,
\newblock ``Multi-source fusion and automatic predictor selection for zero-shot
  video object segmentation,''
\newblock in {\em ACM MM}, 2021, pp. 2645--2653.

\bibitem{dpt}
R.~Ranftl, A.~Bochkovskiy, and V.~Koltun,
\newblock ``Vision transformers for dense prediction,''
\newblock in {\em ICCV}, 2021, pp. 12179--12188.

\bibitem{raft}
Z.~Teed and J.~Deng,
\newblock ``Raft: Recurrent all-pairs field transforms for optical flow,''
\newblock in {\em ECCV}, 2020, pp. 402--419.

\bibitem{resnet}
K.~He, X.~Zhang, S.~Ren, and J.~Sun,
\newblock ``Deep residual learning for image recognition,''
\newblock in {\em CVPR}, 2016, pp. 770--778.

\bibitem{deeplab}
L.-C. Chen, G.~Papandreou, I.~Kokkinos, K.~Murphy, and A.~L. Yuille,
\newblock ``Deeplab: Semantic image segmentation with deep convolutional nets,
  atrous convolution, and fully connected crfs,''
\newblock {\em IEEE TPAMI}, vol. 40, no. 4, pp. 834--848, 2017.

\bibitem{wang2018non}
X.~Wang, R.~Girshick, A.~Gupta, and K.~He,
\newblock ``Non-local neural networks,''
\newblock in {\em CVPR}, 2018, pp. 7794--7803.

\bibitem{hu2018squeeze}
J.~Hu, L.~Shen, and G.~Sun,
\newblock ``Squeeze-and-excitation networks,''
\newblock in {\em CVPR}, 2018, pp. 7132--7141.

\bibitem{2015unet}
O.~Ronneberger, P.~Fischer, and T.~Brox,
\newblock ``U-net: Convolutional networks for biomedical image segmentation,''
\newblock in {\em MICCAI}, 2015, pp. 234--241.

\bibitem{iouloss}
M.~A. Rahman and Y.~Wang,
\newblock ``Optimizing intersection-over-union in deep neural networks for
  image segmentation,''
\newblock in {\em ISVC}, 2016, pp. 234--244.

\bibitem{davis}
F.~Perazzi, J.~Pont-Tuset, B.~McWilliams, L.~Van~Gool, M.~Gross, and
  A.~Sorkine-Hornung,
\newblock ``A benchmark dataset and evaluation methodology for video object
  segmentation,''
\newblock in {\em CVPR}, 2016, pp. 724--732.

\bibitem{fbms}
P.~Ochs, J.~Malik, and T.~Brox,
\newblock ``Segmentation of moving objects by long term video analysis,''
\newblock {\em IEEE TPAMI}, vol. 36, no. 6, pp. 1187--1200, 2013.

\bibitem{segv2}
F.~Li, T.~Kim, A.~Humayun, D.~Tsai, and J.~M. Rehg,
\newblock ``Video segmentation by tracking many figure-ground segments,''
\newblock in {\em ICCV}, 2013, pp. 2192--2199.

\bibitem{vos}
J.~Li, C.~Xia, and X.~Chen,
\newblock ``A benchmark dataset and saliency-guided stacked autoencoders for
  video-based salient object detection,''
\newblock {\em IEEE TIP}, vol. 27, no. 1, pp. 349--364, 2017.

\bibitem{mstm}
W.-C. Tu, S.~He, Q.~Yang, and S.-Y. Chien,
\newblock ``Real-time salient object detection with a minimum spanning tree,''
\newblock in {\em CVPR}, 2016, pp. 2334--2342.

\bibitem{stbp}
T.~Xi, W.~Zhao, H.~Wang, and W.~Lin,
\newblock ``Salient object detection with spatiotemporal background priors for
  video,''
\newblock {\em IEEE TIP}, vol. 26, no. 7, pp. 3425--3436, 2016.

\bibitem{sflr}
C.~Chen, S.~Li, Y.~Wang, H.~Qin, and A.~Hao,
\newblock ``Video saliency detection via spatial-temporal fusion and low-rank
  coherency diffusion,''
\newblock {\em IEEE TIP}, vol. 26, no. 7, pp. 3156--3170, 2017.

\bibitem{scom}
Y.~Chen, W.~Zou, Y.~Tang, X.~Li, C.~Xu, and N.~Komodakis,
\newblock ``Scom: Spatiotemporal constrained optimization for salient object
  detection,''
\newblock {\em IEEE TIP}, vol. 27, no. 7, pp. 3345--3357, 2018.

\bibitem{scnn}
Y.~Tang, W.~Zou, Z.~Jin, Y.~Chen, Y.~Hua, and X.~Li,
\newblock ``Weakly supervised salient object detection with spatiotemporal
  cascade neural networks,''
\newblock {\em TCSVT}, vol. 29, no. 7, pp. 1973--1984, 2018.

\bibitem{pcsa}
Y.~Gu, L.~Wang, Z.~Wang, Y.~Liu, M.-M. Cheng, and S.-P. Lu,
\newblock ``Pyramid constrained self-attention network for fast video salient
  object detection,''
\newblock in {\em AAAI}, 2020, pp. 10869--10876.

\bibitem{smeasure}
D.-P. Fan, M.-M. Cheng, Y.~Liu, T.~Li, and A.~Borji,
\newblock ``Structure-measure: A new way to evaluate foreground maps,''
\newblock in {\em ICCV}, 2017, pp. 4548--4557.

\bibitem{maxf}
R.~Achanta, S.~Hemami, F.~Estrada, and S.~Susstrunk,
\newblock ``Frequency-tuned salient region detection,''
\newblock in {\em CVPR}, 2009, pp. 1597--1604.

\bibitem{mae}
F.~Perazzi, P.~Kr{\"a}henb{\"u}hl, Y.~Pritch, and A.~Hornung,
\newblock ``Saliency filters: Contrast based filtering for salient region
  detection,''
\newblock in {\em CVPR}, 2012, pp. 733--740.

\bibitem{duts}
L.~Wang, H.~Lu, Y.~Wang, M.~Feng, D.~Wang, B.~Yin, and X.~Ruan,
\newblock ``Learning to detect salient objects with image-level supervision,''
\newblock in {\em CVPR}, 2017, pp. 136--145.

\bibitem{monodepth2}
C.~Godard, O.~Mac~Aodha, M.~Firman, and G.~J. Brostow,
\newblock ``Digging into self-supervised monocular depth estimation,''
\newblock in {\em ICCV}, 2019, pp. 3828--3838.

\bibitem{megadepth}
Z.~Li and N.~Snavely,
\newblock ``Megadepth: Learning single-view depth prediction from internet
  photos,''
\newblock in {\em CVPR}, 2018, pp. 2041--2050.

\end{thebibliography}

\end{document}